\documentclass[10pt,twocolumn,letterpaper]{article}

\usepackage{iccv}
\usepackage{times}
\usepackage{graphicx}
\usepackage{amsmath}
\usepackage{amssymb}
\usepackage{algpseudocode}
\usepackage{algorithm}
\usepackage{wrapfig}
\usepackage{pgfplots,ifthen}
\usetikzlibrary{calc}
\usepackage[nomargin,inline,index]{fixme}

\usepackage[pagebackref=true,breaklinks=true,letterpaper=true,colorlinks,bookmarks=false]{hyperref}

 \iccvfinalcopy 

\def\iccvPaperID{615} 

\ificcvfinal
\def\NEW{}
\def\FIX{}
\else
\def\NEW{\color{red}}
\def\FIX{\color{blue}}
\fi

\ificcvfinal\fi

\DeclareMathAlphabet{\mathpzc}{OT1}{pzc}{m}{it}

\begin{document}

\title{Thin Structure Estimation with Curvature Regularization}

\author{Dmitrii Marin, Yuri Boykov, Yuchen Zhong \\
University of Western Ontario, Canada\\
{\tt\small dmitrii.a.marin@gmail.com yuri@csd.uwo.ca yzhong.cs@gmail.com }\\
}

\maketitle

\begin{abstract}
Many applications in vision require estimation of thin structures such as 
boundary edges, surfaces, roads, blood vessels, neurons, etc.
Unlike most previous approaches, we simultaneously detect and delineate 
thin structures with sub-pixel localization and real-valued orientation estimation. 
This is an ill-posed problem that requires regularization. We propose
an objective function combining detection likelihoods with a prior 
minimizing curvature of the center-lines or surfaces.   
Unlike simple block-coordinate descent, we develop a novel algorithm
that is able to perform joint optimization of location and detection variables more effectively.
Our lower bound optimization algorithm applies to quadratic or absolute
curvature.
The proposed early vision framework is sufficiently general
and it can be used in many higher-level applications. 
We illustrate the advantage of our approach on a range of 2D and 3D examples. 
\end{abstract}

\setlength{\intextsep}{2pt}%
\setlength{\columnsep}{8pt}%

\section{Introduction}

A large amount of work in computer vision is devoted to estimation of structures like edges, center-lines, 
or surfaces for fitting thin objects such as intensity boundaries, blood vessels, neural axons, roads, or point clouds.
This paper is focused on the general concept of a {\em center-line}, which could be defined in different ways. 
For example, Canny approach to edge detection implicitly defines a center-line as a ``ridge'' of intensity gradients \cite{canny86}. 
Standard methods for shape skeletons define medial axis as singularities of a distance map from a given object boundary 
\cite{siddiqi2008medial,siddiqi2002hamilton}. In the context of thin objects like edges, 
vessels, etc, we consider a center-line to be a smooth curve
minimizing orthogonal projection errors for the points of the thin structure. 

We study curvature of the center-line as a regularization criteria for its inference. In general,
curvature is actively discussed in the context of thin structures. For example, it is well known that 
curvature of the object boundary has significant effect on the medial axis 
\cite{kimia1995shapes,siddiqi2008medial}. In contrast, we are directly concerned with
curvature of the center-line, not the curvature of the object boundary.
Moreover, we do not assume that the boundary of a thin structure (e.g. vessel or road) is given.
Detection variables are estimated simultaneously with the center-line.
This paper proposes a general energy formulation and an optimization algorithm for detection and subpixel
delineation of thin structures based on curvature regularization. 

\begin{figure}[t]
\includegraphics[clip=true,trim=125mm 2cm 114mm 2cm,width=0.23\textwidth]{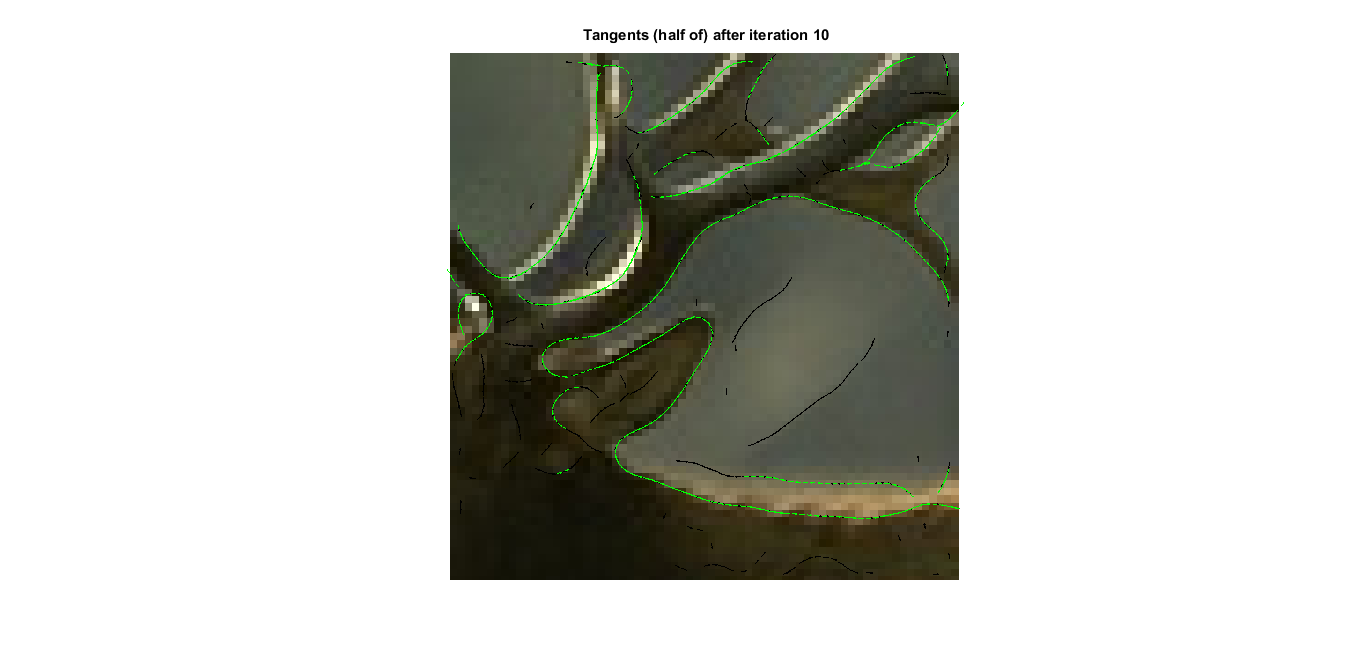}
\includegraphics[clip=true,trim=125mm 2cm 114mm 2cm,width=0.23\textwidth]{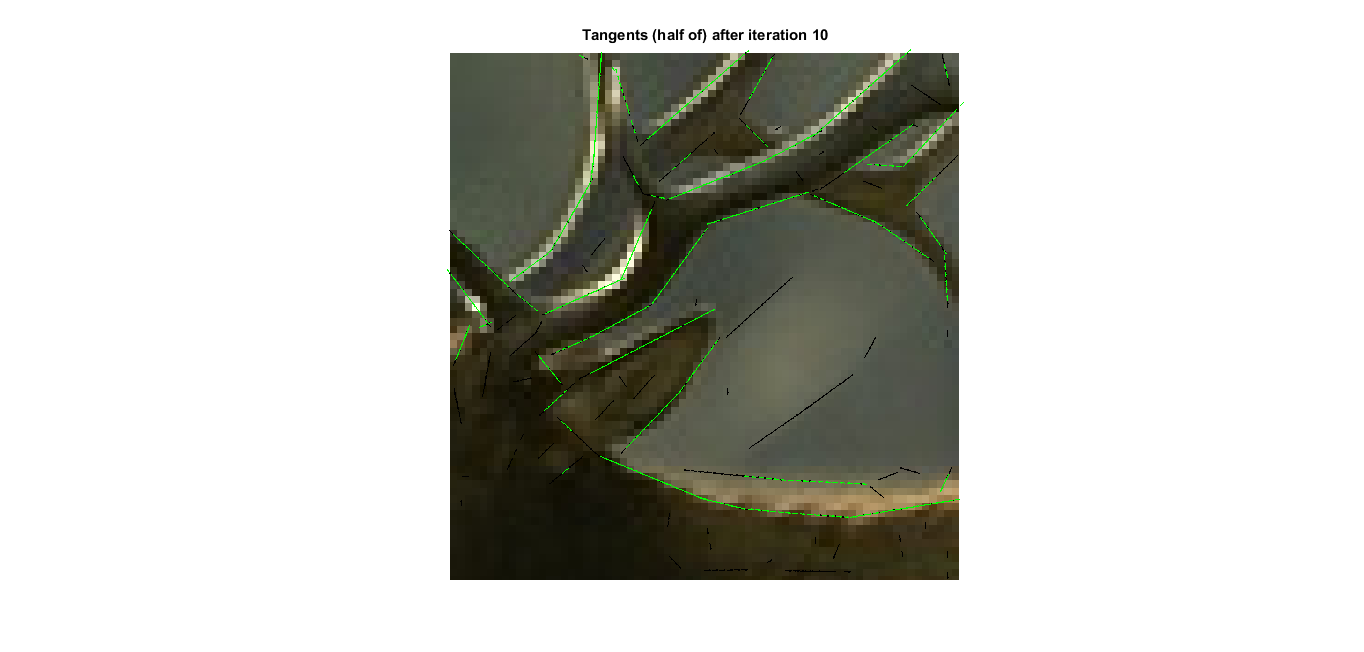}
\caption{{\NEW Edge detection. }
The result of our algorithm for squared (on the left) and
absolute (on the right) curvature approximations. 
{\NEW Green and black lines correspond to edges with
high and medium confidence  measure correspondingly.}
Note the strong bias to straight lines on the right:
the energy prefers a small number of sharp corners 
rather than many smooth corners like on the left.
\label{fig:sq_vs_abs}}
\end{figure}

Curvature is a natural regularizer for thin structures and it has been widely explored in the past.
In the context of image segmentation with second-order smoothness it was studied by
\cite{shoenemann-etal-iccv-2009,strandmark2011curvature,schoenemann-etal-ijcv-2012,Pock:JMIV12,heber-et-al-eccv-2012,olsson2013partial,curvature:cvpr14}.
It is also a popular second-order prior in stereo or multi-view-reconstruction 
\cite{li2010differential, olsson2013defense,woodford2009global}.
Curvature has been used inside connectivity measures for analysis of diffusion MRI
\cite{momayyezsiahkal20133d}. 
Curvature is also widely used for {\em inpainting}
\cite{alvarez1992image,chan2001nontexture} and edge completion 
\cite{guy1993inferring, williams1997stochastic, alter1998extracting}.
For example,  {\em stochastic completion field} technique in
 \cite{williams1997stochastic, momayyezsiahkal20133d} estimates 
probability that a completed/extrapolated curve passes any given 
point assuming it is a random walk with bias to straight paths.
Note that common edge completion methods use existing edge 
detectors as an input for the algorithm. 

In contrast to these prior works, this paper proposes 
a general low-level regularization framework for 
detecting thin structures with accurate estimation of location and orientation. 
In contrast to \cite{williams1997stochastic, guy1993inferring,momayyezsiahkal20133d} 
we explicitly minimize the integral of curvature along
the estimated thin structure. Unlike \cite{guo2014multi} we 
do not use curvature for grouping pre-detected thin structures, 
we use curvature as a regularizer during the detection stage.

\begin{figure*}
\centering
\begin{tabular}{ccc}
\includegraphics[height=5cm]{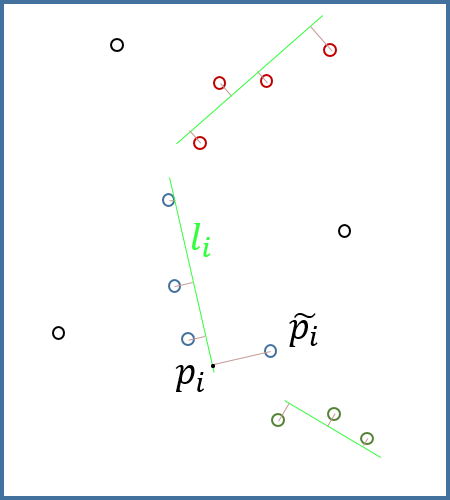} &
\includegraphics[height=5cm]{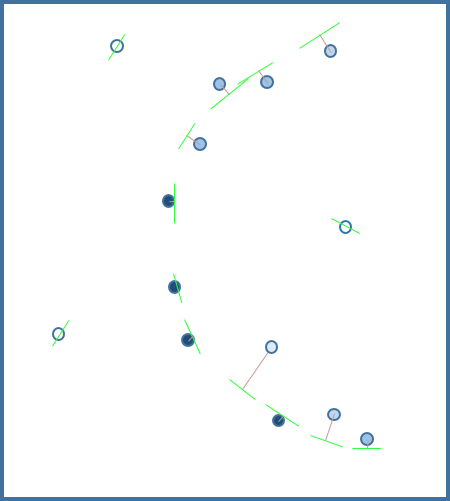} &
\includegraphics[height=5cm]{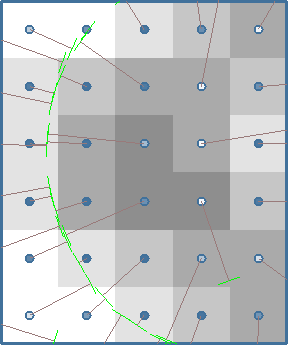} \\
(a) Olsson's model~\cite{olsson2012curvature} & (b) Our model for cloud of points & (c) Our model for grid points
\end{tabular}
\caption{Comparison with~\cite{olsson2012curvature}. {\NEW An empty circle in (b) and (c) denotes low confidence and a dark blue circle means high confidence.}}
\label{pic:cmp-w-carl}
\end{figure*}

{\bf Related work:} 
Our regularization framework is based on the curvature estimation formula proposed 
by Olsson et al.~\cite{olsson2012curvature, olsson2013defense} in the context of
surface fitting to point clouds for multi-view reconstruction, see 
Fig.\ref{pic:cmp-w-carl}(a). One assumption in~\cite{olsson2012curvature,
olsson2013defense} is that the data points are noisy readings
 of the surface. While the method allows outliers,
their formulation is focused on estimation of local surface patches. Our work can be
 seen as a generalization to detection problems where majority of the data points, 
e.g. image pixels in Fig.\ref{pic:cmp-w-carl}(c), are not within a thin structure. 
In addition to local tangents, our method estimates probability that the point is a 
part of the thin structure. Section~\ref{sec:energy} discusses in details this and 
other significant differences from the formulation in~\cite{olsson2012curvature, olsson2013defense}.

\begin{wrapfigure}{r}{34mm}
\includegraphics[width=34mm]{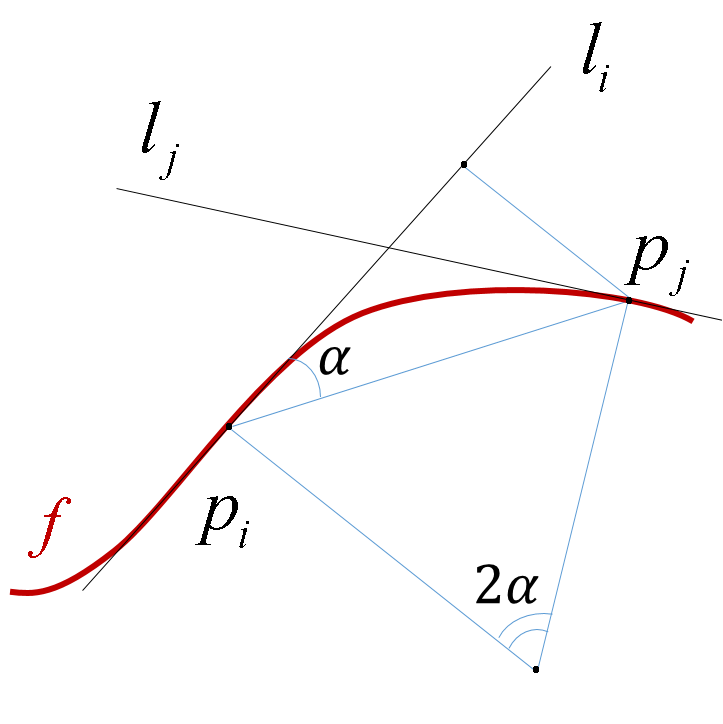}
\end{wrapfigure} 

Assuming $p_i$ and $p_j$ are neighboring points on a thin structure, \eg a curve,
 Olsson et al. \cite{olsson2012curvature} 
evaluate local curvature as follows. Let $l_i$ and $l_j$ be the tangents to the
curve at points $p_i$ and $p_j$. Then the authors propose the following
approximation for the absolute curvature 
\begin{equation*}
|\kappa(l_i,l_j)|  = \frac{||l_i-p_j||+||l_j-p_i||}{||p_i-p_j||} 
\end{equation*}
and for the squared curvature 
\begin{equation*}
\kappa^2(l_i,l_j) = \frac{||l_i-p_j||^2+||l_j-p_i||^2}{||p_i-p_j||^2} \label{eq:sq-curv}
\end{equation*}
where $||l_i-p_j||$ is the distance between point $p_j$ 
and line $l_i$.

Assume that the curve \mbox{$\vec r = f(\tau)$} is parameterized by arc-length $\tau$ such that
$\tau_1\le \tau\le \tau_M$. If $(\tau_1,\tau_2,  \dots,\tau_M)$ is 
an increasing parameter sequence then the curvature of $f$ can be approximated by
$$\int|\kappa|^\alpha d\tau \;\;\approx \sum_{(i,j)\in N} |\kappa(l_i,l_j)|^\alpha$$
where $N=\{(i, i+1)\ |\ i=1,2,\dots M-1\}$ is a neighborhood system for
curve points $p_i=f(\tau_i)$ and $l_{i}=\dot f(\tau_i)$ are their tangent lines.

Olsson et al.~\cite{olsson2012curvature} use regularization for fitting a surface
(or curve) to a cloud of points in 3D (or 2D) space.
Every observed point $\tilde{p_i}$ is treated as a noisy measurement of some unknown point $p_i$ that is
the closest point on the estimated surface,  see Fig.\ref{pic:cmp-w-carl}(a). 
Each $\tilde p_i$ is associated with unknown local surface patch $l_i$ that is 
a tangent plane for the surface at $p_i$. The proposed surface fitting energy combines 
curvature-based regularization with the first order data fidelity term
\begin{equation}
E(L) = \sum_{(i,j)\in N} |\kappa(l_i,l_j)|^{\alpha}w_{ij} + \sum_{i}\frac1{\sigma^2}||l_i -\tilde p_i||^2 
\label{eq:energy-carl}
\end{equation}
where $L=\{l_i\}$ is the set of tangents, 
$N$ is a neighborhood system,  
$\sigma$ is non-negative constant, $w_{ij}$ is a positive constant 
such that $\sum_{j\in N_i}w_{ij}=1$. To minimize~\eqref{eq:energy-carl}, 
the algorithm in~\cite{olsson2012curvature} iteratively optimizes the assignment 
variables for a limited number of tangent proposals, and then re-estimates tangent 
	plane parameters, see Fig.\ref{pic:cmp-w-carl}(a).

In contrast to \cite{olsson2012curvature}, our method estimates thin structures 
in the image grid where, {\em a priori}, it is unknown which pixels belong to the thin structure, see
Fig.\ref{pic:cmp-w-carl}(c). We introduce set $X=\{x_i\}$ of indicator variables $x_i \in \{0,1\}$ 
where $x_i=1$ iff pixel $\tilde {p}_i$ belongs to the thin structure. Our basic energy~\eqref{eq:energy-w-x}
and its extensions combine unary detection potentials with curvature regularization. Due to the regularity of
our grid neighborhood, we use constant weights $w_{ij}$, which are omitted from now on. 
We propose a different optimization technique estimating a posteriori
distribution of $x_i$ and separate tangents $l_i$ at each point. As illustrated in Fig.\ref{pic:cmp-w-carl}(b), 
our framework is also applicable to energy~\eqref{eq:energy-carl} and multi-view reconstruction problem as in
\cite{olsson2012curvature,olsson2013defense}.

Parent\&Zucker~\cite{zucker:89} formulate a closely related {\em trace inference} problem for detecting curves in
2D grid. Similarly to us, they estimate indicator variables $x_i$ and tangents $l_i$.  
However, they estimate  $x_i$ and $l_i$ by enforcing a {\em co-circularity} constraint 
assuming given local {\em curvature information}, which they estimate in advance.
In contrast, we simultaneously estimate $x_i$ and $l_i$ by optimizing objective~\eqref{eq:energy-w-x}
that directly regularizes curvature of the underlying thin structure.
Moreover, \cite{zucker:89} quantizes curvature information and tangents 
while our model uses real valued curvature and tangents. The extension of~\cite{zucker:89} 
to 3D is not trivial.

{\NEW Similarly to \cite{olsson2012curvature,zucker:89} we estimate tangents only at a \emph{finite} set of points. 
Additional regularization is required if continuous center-line between these points is needed \cite{kamberov2000ill}.}

{\bf Contributions:} It is known that curvature of an object boundary is an important shape descriptor \cite{sebastian2004recognition} with a significant effect on {\em medial axis} \cite{kimia1995shapes,siddiqi2008medial},
which is not robust even to minor perturbations of the boundary. In the context of thin objects 
(e.g. edges, vessels) we study a concept of a center-line (a smooth 1D curve minimizing the sum of projection errors),
which is different from {\em medial axis}. We regularize the curvature of the center-line.
Unlike many standard methods for center-lines, we do not assume that the shape of the object is given and
propose a general low-level vision framework for thin structure detection combined with
sub-pixel localization and real-valued orientation of its center-line. 
Therefore, we propose an approach that takes
into account all possible configurations of the indicator variables while estimating the tangents.
This significantly improves stability with respect to local minima.
Our optimization method uses variational inference and trust region frameworks
adapted to absolute and quadratic curvature regularization.


Our proof-of-the-concept experiments demonstrate encouraging results
in the context of edge and vessel detection in 2D and 3D images. 
In particular, we obtain promising results for estimating highly detailed
vessels structure on high-resolution microscopy CT volumes.
We also show examples of sub-pixel edge detection regularizing curvature.
While there are no databases for comparing edge detectors with real-valued
location and orientation estimation, we obtained competitive results on a pixel-level
edge detection benchmark \cite{guo2012evaluating}.
Our general early vision methodology can be integrated into higher semantic level boundary 
detection techniques, e.g. \cite{BSDS}, but this is outside the scope of this work.
Our current sequential implementation is not tuned to optimize performance.
Its running time for edges in 2D image of Fig.\ref{fig:sq_vs_abs} is 20 seconds and for vessels
in 3D volume of Fig.\ref{pic:vessels1} is one day. However, our method is highly-parallelizable on GPU
and fast real-time performance on 2D images can be achieved. 

In Section~\ref{sec:energy} we describe the proposed model and 
discuss a simple block-coordinate descent 
optimization algorithm and its drawbacks. In Section~\ref{sec:opt} we 
propose a new optimization method for our energy based on variational inference framework.
In Section~\ref{sec:tr} we describe the details of the proposed method and discuss the difference between squared 
and absolute curvatures (Subsection \ref{sec:sqabs}). We describe several applications of the 
proposed framework in Section~\ref{sec:application} and conclude in Section~\ref{sec:conclusion}.

\section{Energy formulation} \label{sec:energy}

In the introduction we informally defined the center-line of a thin structure as a smooth curve minimizing
orthogonal projection errors. Here we present the energy formalizing this
criterion. First we note that in our model the curve is not defined explicitly but through
points $p_i$ it passes and tangent lines $l_i$ at these points. The energy is given by
\begin{align}
E\ &(L,X) = \sum_{(i,j)\in N}  \kappa^2(l_i,l_j)x_ix_j + \notag\\
& + \sum_{i} \frac1{\sigma^2}||l_i - \tilde p_i||_+^2x_i + {\NEW \sum_i}\lambda_i x_i \label{eq:energy-w-x}
\end{align} 
where $N$ is a neighborhood system, 
{\NEW$X=\{x_i\}$ is a set of indicator variables $x_i \in \{0,1\}$ 
where $x_i=1$ iff pixel $\tilde {p}_i$ belongs to the thin structure,}
$\lambda_i$ define unary potentials penalizing/rewarding presence 
of the structure at $\tilde p_i$. In contrast to~\eqref{eq:energy-carl},
potentials $\lambda_i$ define the data term
 while $\frac1{\sigma^2}||l_i -\tilde p_i||_+^2$ is a soft constraint.

\definecolor{rose}{rgb}{0.7813,0.5938,0.5938}

\begin{wrapfigure}{r}{5mm}
\begin{tikzpicture}[scale=0.3,dot/.style={circle,inner sep=0.1pt,fill},]
  \coordinate (A) at  (0,0) ;
  \coordinate (B) at  (2,12) ;
  \foreach \i in {0,...,3} {
     \foreach [evaluate={\d=int(abs(-6*\i/sqrt(37)+\j/sqrt(37)))}]  \j in {1,...,11} {
        \ifthenelse{\d<1}{
           \draw[-,color=rose] (\i,\j)  -- ($(A)!(\i,\j)!(B)$);
           \node  [dot] at (\i,\j) {};}{};
     }
  }
  \draw[-,color=green] (A) -- (B);
  
\end{tikzpicture}
\end{wrapfigure}

We explore two choices of the soft constraint \mbox{$||l_i-\tilde p_i||_+$}. 
The first one uses Euclidean distance. 
In that case it models normally distributed errors. 
Although it is appropriate for many applications, 
e.g. surface estimation in multi-view reconstruction~\cite{olsson2012curvature,olsson2013defense}, 
the normal errors assumption is no longer valid for the image grid 
because the discretization errors are not Gaussian.
In fact, using Euclidean distance may make the 
soft constraint term proportional to the length of the center-line{\NEW, see illustration on the right}.

\begin{wrapfigure}{r}{28mm}
\begin{tikzpicture}[scale=0.5]
  \draw[->] (-2,0) -- (2,0) node[right] {$d$};
  \draw[-] (0,-0.2) -- (0,0.9) node[above] {};
  \draw[thick,scale=1,domain=-2:2,smooth,variable=\x,blue] plot ({\x},{max(0,abs(\x)-1)^2});
  \node[draw,align=center,blue,draw=none] at (0.2,1.5) {$\max(0, |d| - 1)^2$};
\end{tikzpicture}
\end{wrapfigure}

Thus, we also propose a truncated form of Euclidean distance: 
\begin{equation}||l_i - \tilde p_i||_+=\max(0, ||l_i - \tilde p_i|| - 1).\label{eq:disc-error}\end{equation}
This does not penalize tangent lines $l_i$ that are within 
one pixel from points $\tilde p_i$. Different applications may require a 
different choice of no-penalty threshold. 

\paragraph{Extensions.}

We can extend the energy~\eqref{eq:energy-w-x} by adding other terms 
that encourage various other useful properties. For example, energy
\begin{equation}
E'(L,X) = E(L,X)-\gamma \sum_{(i,j)\in N}x_ix_j
\label{eq:energy-w-gamma}
\end{equation}
for $\gamma>0$ will reward well aligned tangents. The effect of this term is 
shown in Fig.\ref{pic:gamma-effect}. This term is similar to edge ``repulsion'' in 
MRF-based segmentation. The overall pairwise potential $(\kappa(l_i,l_j)-\gamma)x_ix_j$ 
encourages edge continuity.

Another extension is to use prior knowledge about the center-line direction 
$g_i$ at pixel $\tilde p_i$:
\begin{equation}
E'(L,X) = E(L,X) +\beta\sum_i{\NEW m}(l_i, g_i)^2x_i.
\label{eq:energy-w-beta}
\end{equation}
The term ${\NEW m}(l_i, g_i)$ measures how well tangent line $l_i$ is aligned with prior $g_i$:
$${\NEW m}(l_i, g_i) = ||g_i||\sin\angle(l_i,g_i). $$
The magnitude of $g_i$ constitutes the confidence measure. For example, vectors $g_i$ could be 
obtained from the image gradients or the eigenvectors in the vesselness measure~\cite{frangi1998multiscale}.

\subsection{Block-coordinate descent optimization}
\label{sec:bc}
To motivate our optimization approach for energy~\eqref{eq:energy-w-x} 
described in Section~\ref{sec:vi}, first we describe a simpler optimization 
algorithm and discuss its drawbacks.

The most obvious way to optimize energy \eqref{eq:energy-w-x} 
is a block-coordinate descent. The optimization alternates two steps 
described in Alg.\ref{al:bc}.
The auxiliary energy optimized on line \ref{al:bc:opt-l} is 
a non-linear least square problem and can be optimized by 
a trust-region approach, see Section \ref{sec:tr}. 
The auxiliary function on line \ref{al:bc:opt-x} is a non-submodular 
binary pairwise energy that can be optimized with TRWS\cite{trws}.

\begin{algorithm}
\caption{Block-coordinate descent}\label{al:bc}
\begin{algorithmic}[1]
\State Initialize $L^0$ and $X^0$
\State $k\leftarrow0$
\While {not converged}
    \State Optimize $L^{k+1} \leftarrow \arg\min_L E(L, X^k)$    \label{al:bc:opt-l}
    \State Optimize $X^{k+1} \leftarrow \arg\min_X E(L^{k+1}, X)$ \label{al:bc:opt-x}
    \State $k\leftarrow k+1$
\EndWhile
\end{algorithmic}
\end{algorithm}

\begin{figure}[t]\centering
\includegraphics[trim=13cm 10cm 13cm 11cm,clip=true,width=0.47\textwidth]{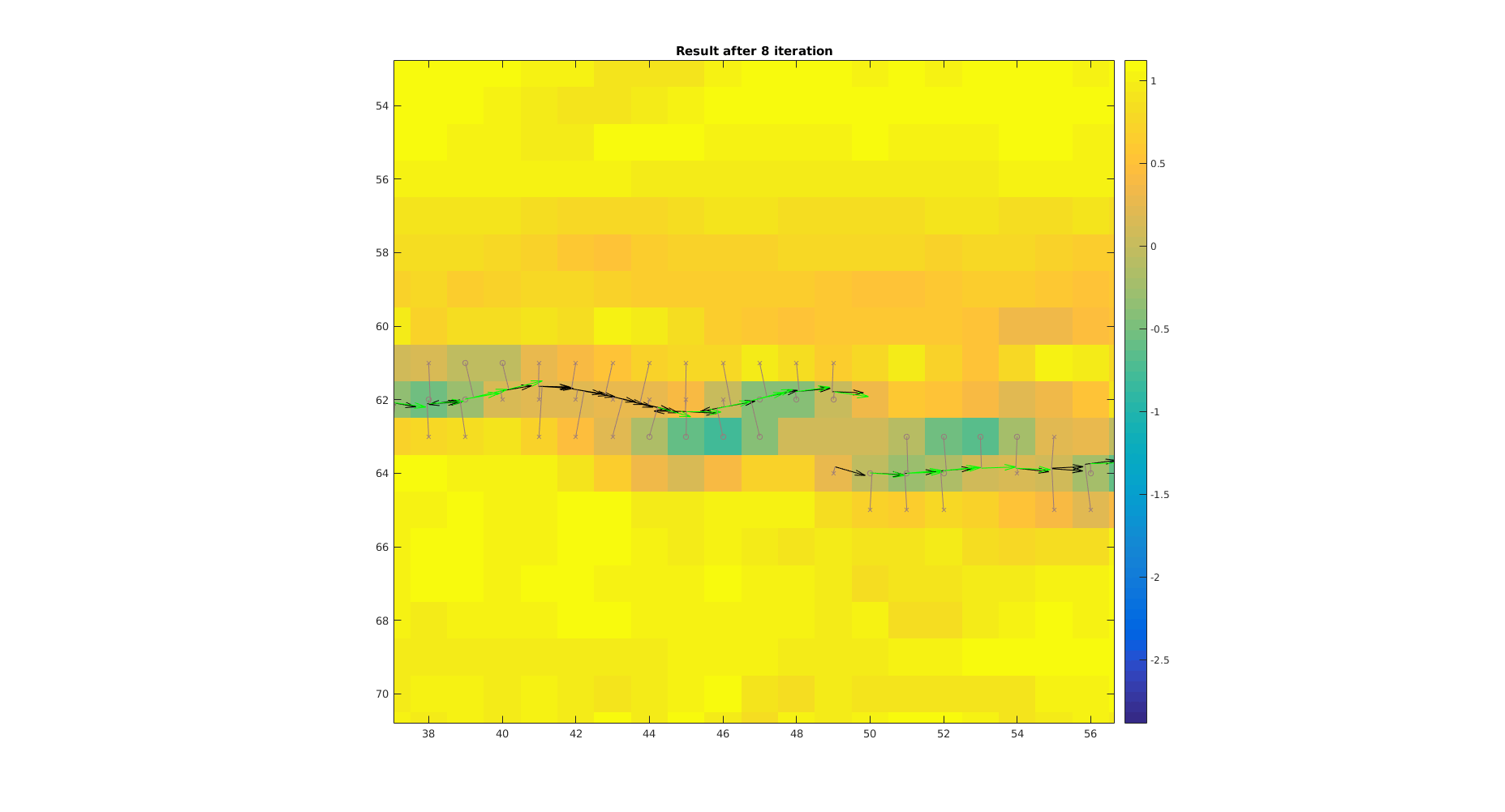} 
\caption{An example of local minima for Alg.\ref{al:bc}. 
The more ``blue'' is a pixel, the more likely it is to lie on an edge. 
Green arrows correspond to pixels that were initialized as edges. 
Black arrows correspond to the  edges detected by Alg.\ref{al:bc}. 
This local minimum consists of two disconnected center-lines. 
The globally minimum solution smoothly connects the two pieces into a single center-line.
}\label{fig:bc-heu}
\end{figure}

We found that Alg.\ref{al:bc} is extremely sensitive to local minima, see Fig.\ref{fig:bc-heu}.
The reason is that tangents $l_i$ for points with indicator variables 
$x_i^k=0$ do not participate in optimization on line~\ref{al:bc:opt-l}. To improve performance of block-coordinate descent, 
we tried heuristics to extrapolate tangents into such regions. We found that good heuristics should have the following properties:

\begin{wrapfigure}{r}{25mm}
\begin{tikzpicture}[scale=0.5]
  \draw[thick,domain=0:360,smooth,variable=\x,blue] plot ({cos(\x)+0.5},{sin(\x)});
  \node[draw,align=center,blue,draw=none] at (0.5,-1.5) {$2\pi$};
  \draw[thick,domain=0:360,smooth,variable=\x,brown] plot ({(cos(\x)*(1-\x/1000)+3.5)},{sin(\x)*(1-\x/1000)});
  \draw[thick,-,brown] (4.5-0.36,0) -- (4.5,0) ;
  \node[draw,align=center,brown,draw=none] at (3.5,-1.5) {$2\pi+\pi$};
\end{tikzpicture}
\end{wrapfigure}

1. Since {\NEW integral of curvature is sensitive to small local errors 
(see the figure on the right),}
the extrapolating procedure should yield close tangents
for neighbors. Otherwise step \ref{al:bc:opt-x} of the algorithm
is ineffective.  This issue could be partially solved
by using energy \eqref{eq:energy-w-gamma}. In this case it can be  
beneficial to connect two tangents even if there is some misalignment error.

2. The heuristic should envision that some currently disconnected
curves may lie on the same center-line, see Fig.\ref{fig:bc-heu}.

The first property was easy to incorporate, while the second would require sophisticated edge continuation 
methods, e.g. a stochastic completion field~\cite{williams1997stochastic, momayyezsiahkal20133d}.
Instead we develop a new optimization procedure (Section~\ref{sec:vi}) based on variational inference. 
The advantage of our new procedure is that it is closer to joint optimization of $L$ and $X$.


\section{Variational Inference}\label{sec:opt}\label{sec:em-details}\label{sec:vi}

Ideally, we wish to jointly optimize~\eqref{eq:energy-w-x} with respect to all variables. 
This is a mixed integer non-linear problem with an enormous number of variables. Thus, it is intractable.
However, we can introduce elements of joint optimization based on stochastic \emph{variational 
inference} framework. The proposed approach takes into account all possible configurations 
of indicator variables $x_i$ while estimating tangents $l_i$.
This significantly improves stability w.r.t. local minima.


Energy \eqref{eq:energy-w-x} corresponds to a Gibbs distribution:
\begin{equation*}
P(I,X,L') = \frac 1Z \exp \left( -E(L',X) \right)
\end{equation*}
where $Z$ is a normalization constant and the image is given by 
data fidelity terms $I=\{ \lambda_i \}$. Here $I$ are visible 
variables, indicator variables $X$ and tangents $L'=\{l'_i\}$ are hidden ones. 
{\NEW We add a prime sign for tangent notation to distinguish values of random variables 
and parameters of the distribution.} Our goal is to 
approximate the posterior distribution $P(X,L'|I)$ of unobserved (hidden) indicators $X$ 
and tangents $L'$ given image $I$. The problem of approximating the posterior distribution 
has been extensively studied and is known as \emph{variational inference}~\cite{bishop2006pattern}.

Variational inference is based on the decomposition
\begin{equation}
\ln P(I)=\mathcal L(q) + \operatorname {KL}(q||p)
\label{eq:decomp}
\end{equation}
where $\ln P(I)$ is the \emph{evidence}, $q(X,L')$ is a distribution over 
the hidden variables,  $p(X,L')=P(X,L'|I)$ is the posterior distribution, and
\begin{align}
\mathcal L(q) &= \sum_X \int q(X,L') \ln \left( \frac{P(I,X,L')}{q(X,L')} \right)dL',\label{eq:lower-bound}\\
\operatorname{KL}(q||p) &= -\sum_X \int q(X,L') \ln \left( \frac{P(X,L'|I)}{q(X,L')}\right) dL'.
\end{align}

Since $\operatorname{KL}$ (Kullback--Leibler divergence) is always non-negative, 
the functional $\mathcal L(q)$  is a lower bound for the evidence $\ln P(I)$. 
One of the nice properties of this decomposition is that the global maximum 
of lower bound $\mathcal L$ coincides with the global minimum of  $\operatorname{KL}(q||p)$
 and optimal $q^*(X,L')=\arg\max_q\mathcal L(q)$ is equal to the true posterior 
$P(X,L'|I)$~\cite{bishop2006pattern}.

Unfortunately~\eqref{eq:lower-bound} cannot be optimized exactly. 
To make optimization tractable, in variational inference framework 
one assumes that $q$ belongs to a family of suitable distributions. 
In this work we will assume that $q$ is a factorized distribution
 (\emph{mean field theory}~\cite{parisi1988meanfield}):
\begin{align}
q(X,L')=&q(X)q(L'),\\
q(X)=&\prod_{i}q_i(x_i)=\prod_{i}q_i^{x_i}(1-q_i)^{1-x_i},\\
q(L')=&\prod_{i}\delta(l'_i-l_i)\label{eq:iid-x}
\end{align}
where {\NEW $\delta(l'_i-l_i)$ is a deterministic (degenerate) distribution with parameter $l_i$}. 
Under this assumption lower bound functional $\mathcal L$ becomes a function of 
parameters $q_i$ and $l_i$. We denote this function $\mathcal L(Q, L)$ where $Q=\{q_i\}$ and $L=\{l_i\}$.

The proposed algorithm is defined by Alg.\ref{al:vi}. It optimizes 
lower bound $\mathcal L(Q,L)$ in block-coordinate fashion. 
The algorithm returns optimal tangents $l_i^*$, see Fig.\ref{pic:result-edges}(b), 
and optimal probabilities $q_i^*$, see Fig.\ref{pic:result-edges}(c).

\begin{algorithm}
\caption{Block-Coordinate Descend for Variational Inference}\label{al:vi}
\begin{algorithmic}[1]
\State Initialize $L^0$ and $Q^0$ \label{al:em:init}
\State $k\leftarrow0$
\While {not converged}
    \State Optimize $L^{k+1} \leftarrow \arg\max_L \mathcal L(Q^k,L)$    \label{al:em:opt-l}
    \State Optimize $Q^{k+1} \leftarrow \arg\max_Q \mathcal L(Q,L^{k+1})$ \label{al:em:opt-x}
    \State $k\leftarrow k+1$
\EndWhile\\
\Return $L^{k}$, $Q^{k}$
\end{algorithmic}
\end{algorithm}

Now we consider optimization of $\mathcal L$ over $L$. 
Taking into account \eqref{eq:iid-x}, \eqref{eq:energy-w-x} 
and \eqref{eq:lower-bound} we can derive
\begin{align}
 \arg& \max_L\mathcal L(Q^k,L) 
= \arg\min_L \sum_X q^k(X) E(X,L) = \notag \\
= &\arg\min_L  \sum_{(i,j)\in N} \psi_{ij}q^k_iq^k_j + \sum_{i}\psi_i q^k_i. \label{eq:e-energy}
\end{align}
where
\begin{align*}
\psi_{ij} &\equiv \kappa^2(l_i,l_j), \\
\psi_i &\equiv \frac1{\sigma^2}||l_i - \tilde p_i||_+^2 + \lambda_i.
\end{align*}
{\NEW In case of \eqref{eq:energy-w-gamma} we redefine 
$\psi_{ij}\equiv \kappa^2(l_i,l_j)-\gamma$, and 
in case of \eqref{eq:energy-w-beta} we redefine 
$\psi_{i}\equiv \frac1{\sigma^2}||l_i - \tilde p_i||_+^2 + \lambda_i + \beta m(l_i,g_i)$.}

We see that optimization of $\mathcal L(Q^k,L)$  with respect to $L$ is a non-linear least 
square problem. For optimization details please refer to Section~\ref{sec:tr}.

The optimization w.r.t. $Q$ can be done by coordinate descent. The update equation is~\cite{bishop2006pattern}
\begin{align}\ln q_i^*(x_i)&=\operatorname{E}_{j\ne i}[\ln P(I,X|L)] + const=\notag \\
&= -x_i\left(\sum_{j:(i,j)\in N}\psi_{ij}q_j+\psi_i\right)+const. \label{eq:q-update}
\end{align}

\fxnote{How should we treat $\beta$ and $\gamma$ in this equations?}

The constant in expression \eqref{eq:q-update} does not depend on $x_i$ and 
thus can be determined from the normalization equation $q_i(1)+q_i(0)=1.$
We initialize $q^0(x_i) = \exp(-x_i \psi_i) / (1 + \exp(-\psi_i))$ on line 
\ref{al:em:init} of Alg.\ref{al:vi}. We iterate over all pixels
update step \eqref{eq:q-update} on line \ref{al:em:opt-x} until convergence, 
which is guaranteed by convexity of $\mathcal L$ with respect to each 
$q_i$~\cite{bishop2006pattern}.

Note that if we further restrict $q$ to be a degenerate distribution
(meaning $q(x_i) \in \{0,1\}$) we will get the block-coordinate 
descend Alg.\ref{al:bc}.

The initialization of $L^0$ is application dependent. In many cases some
information about direction of a thin structure is available. 
Concrete initialization examples are described in Section~\ref{sec:application}.

\paragraph{Alternative interpretations.}
The goal of Alg.\ref{al:bc} is to find $\min_{L,X} E(L,X)$, which is equivalent to
\begin{equation}\max_L\max_X (-E(L,X)).\label{eq:max-min-energy}\end{equation}
As shown in Section~\ref{sec:energy} optimization of~\ref{eq:max-min-energy}
in a block-coordinate fashion requires optimization 
of tangents $L$ with fixed indicator variables $X$. This necessitates extrapolation of tangents. 
Instead we propose to optimize $L$ taking into account all possible configurations of $X$. 
That is we propose to replace maximum with smooth maximum:
$$\max_L\sum_X \exp(- E(L,X)).$$
Then we can write down a decomposition similar to~\eqref{eq:decomp}, 
which provides a lower bound yielding the same optimization procedure.

The proposed procedure is closely related to the EM algorithm\cite{1977EM}
where we treat tangents $L$ as the parameters of the distribution. However, in this case 
the normalization constant of the distribution depends on $L$ and 
optimization problem is intractable. One possible way to fix this issue 
is to use a \emph{pseudo likelihood}~\cite{li2009markov}.


\section{Trust region for tangent estimation}\label{sec:tr}

Optimization of the auxiliary functions on line~\ref{al:bc:opt-l} of Algorithms~\ref{al:bc} 
and~\ref{al:vi} as well as energy~\eqref{eq:energy-carl}
is a non-linear least square problem. 
In~\cite{olsson2012curvature,olsson2013defense} 
energy~\eqref{eq:energy-carl} is optimized using discrete multi-label 
approach in the context of surface approximation. 
In our work we adopt the inexact Levenberg-Marquardt method in~\cite{wright1985inexact}, 
which is a trust region second order continuous iterative optimization method.

Each iteration consists of several steps. First, the method linearizes:
\begin{align*}
\mathcal L(q^k&,L+\delta L)\approx \mathcal L(\delta L) \equiv\\
\equiv& \sum_{(i,j)\in N}\left (|\kappa(l_i,l_j)| + \frac{\partial\kappa}{\partial l_i}\delta l_i +  \frac{\partial\kappa}{\partial l_j}\delta l_j \right)^2q^k_iq^k_j +  \\
& + \sum_{i}\frac1{\sigma^2}\left(||l_i - \tilde p_i||_+ +\frac{\partial d}{\partial l_i}\delta l_i \right)^2q^k_i 
\end{align*}
where for compact notation we define 
$\kappa\equiv |\kappa(l_i,l_i)|$ and $d\equiv||l_i - \tilde p_i||_+$. 
We use~\cite{ceres-solver} for automatic calculation of derivatives.

Second, the algorithm solves the minimization problem
$$\delta L^*=\arg\min_{\delta L}\mathcal L(\delta L) + \lambda ||\delta L||^2$$
where $\lambda$ is a positive damping factor, which determines the trust region.
The method uses an inexact iterative algorithm for this task.

The last stage of iteration is to compare the predicted 
energy change $\mathcal L(\delta L^*)-\mathcal L(\vec0)$ with the 
actual energy change $\mathcal L(q^k,L+\delta L^*)-\mathcal L(\vec0)$. 
Depending on the result of comparison the method updates 
variables $L$ and damping factor $\lambda$.
For more details please refer to~\cite{wright1985inexact}.

The most computationally expensive part of Alg.\ref{al:vi} is
trust region optimization described in this subsection. From the technical point of view it consists 
of derivatives computation and basic linear algebra operations. Fortunately, these operations could be easily 
parallelized on GPU. We leave the GPU implementation for a future work.

\subsection{Quadratic vs Absolute Curvature}\label{sec:sqabs}

Previous sections assume squared curvature, but everything can be 
adapted to the absolute curvature too. We only need to discuss how to optimize 
\eqref{eq:e-energy} for the absolute curvature. 
We use the following approximation:
\begin{equation}\label{eq:abs-aprox}
{\NEW \frac{||l_i-p_j||}{||p_i-p_j||} \approx \frac{||l_i-p_j||^2}{||p_i-p_j||^2} \cdot w_{ij}}
\end{equation}
where 
$$ w_{ij} =  \frac{||p_i-p_j||+\epsilon}{||l_i-p_j||+\epsilon} $$
and $\epsilon$ is some non-negative constant. 
If $\epsilon=0$ we have an approximation 
of the absolute curvature, if $\epsilon\to\infty$ we have 
an approximation of the squared curvature.

The trust region approach (see Section \ref{sec:tr}) 
works with approximations of functions.
It does not require any particular approximation 
like in the Levenberg-Marquardt method~\cite{levenberg1944,wright1985inexact}. 
Thus we can approximate the absolute curvature by treating 
$w_{ij}$ as constants in \eqref{eq:abs-aprox} and 
linearizing $\kappa(l_i,l_j)$ analogously to the squared curvature case.

\begin{wrapfigure}{r}{25mm}
\begin{tikzpicture}
  \draw[->] (-0.2,0) -- (1.6,0) node[right] {$\alpha$};
  \draw[->] (0,-0.2) -- (0,1.6) node[above] {$\kappa$};
  \draw[thick,domain=0:1.57,smooth,variable=\x,blue] plot ({\x},{abs(\x)});
  \draw[thick,domain=0:1.57,smooth,variable=\x,red] plot ({\x},{sin(\x/3.14*180)});
  \node[draw,align=left,red,draw=none] at (1,0.2) {$\sin\alpha$};
  \node[draw,align=left,blue,draw=none] at (1,1.2) {$\alpha$};
\end{tikzpicture}
\end{wrapfigure}
The approximation of curvature given by~\cite{olsson2012curvature} 
is derived under the assumption that the angles between neighbor 
tangents are small. Under this assumption the sine of an angle 
is approximately equal to the angle. And the approximation essentially 
computes the sines of the angles rather than the angles themselves.
As a result it significantly underestimates the curvature of sharp corners.

\begin{figure}[t]
\includegraphics[clip=true,trim=2cm 35mm 6cm 3cm,width=0.23\textwidth]{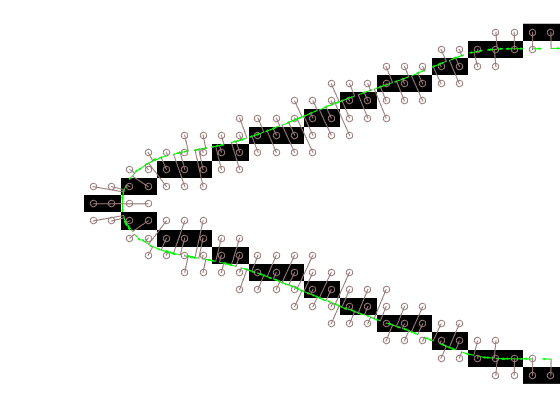}
\includegraphics[clip=true,trim=2cm 35mm 6cm 3cm,width=0.23\textwidth]{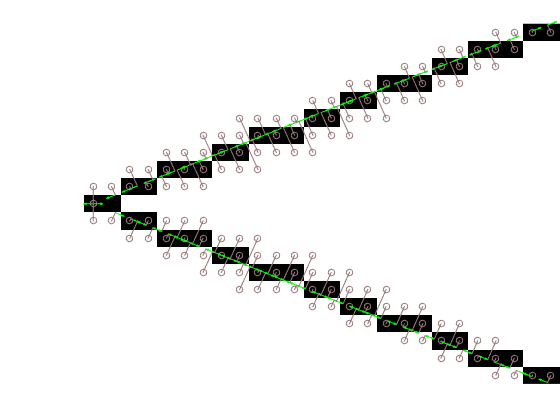}
\caption{The difference between squared (left) and
absolute (right) curvature approximations on an artificial example. 
Note the ballooning bias of squared curvature.}\label{fig:sq_vs_abs_art}
\end{figure}

For example, let us consider the integral of absolute curvature over a circle 
and a square. The integral of the approximation is $2\pi$ and $4$ 
correspondingly, while the integral of the true absolute curvature is 
$2\pi$ in both cases. So the energy using this approximation of absolute
curvature tends to distribute curvature into a small number of sharp 
corners showing strong bias to straight lines. Although approximation 
of squared curvature also underestimates curvature of sharp 
corners, it does not have a strong bias to straight lines.
See figures \ref{fig:sq_vs_abs} and \ref{fig:sq_vs_abs_art} for comparison of
the approximations.

\section{Applications}\label{sec:application}

\subsection{Contrast edges}\label{sec:edges}

\begin{figure*}[t]
\begin{tabular}{cccc}
\includegraphics[height=35mm]{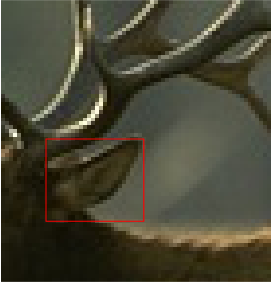} &
\includegraphics[height=35mm]{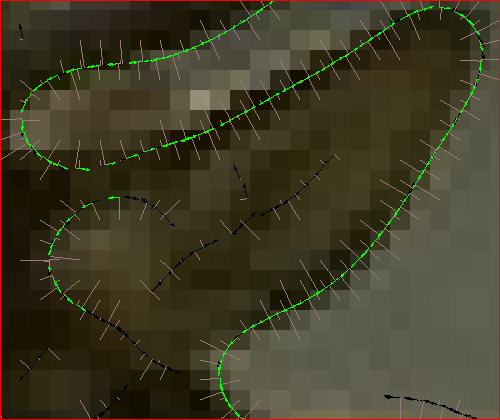} &
\includegraphics[height=35mm]{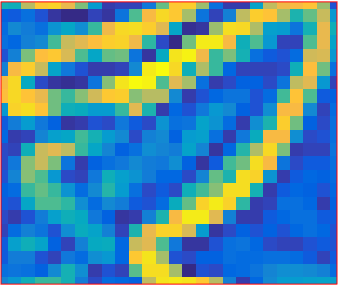} &
\includegraphics[height=35mm]{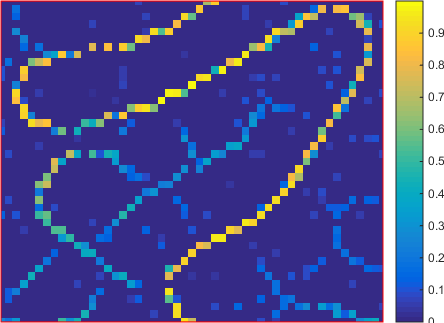} \\
(a) Original image & (b) The output of algorithm & (c) Probabilities $q_i$ & (d) Subpixel probabilities $\tilde q_{p}$
\end{tabular}
\caption{The result of the proposed algorithm. The original image is shown on (a). 
The zoomed in region is shown with a red box. Estimated tangents are shown in (b). 
Green color denotes tangents corresponding to pixels $\tilde p_i$ such that $q_i \ge \frac12$, 
and tangents corresponding to pixels with $q_i\ge\frac14$ are shown in black. 
(c) shows probabilities $q_i$. (d) shows the probabilities at doubled 
resolution produced by projecting points to theirs tangents:  $\tilde q_{p}=q_i$.}
\label{pic:result-edges}
\end{figure*}

Here we consider an application of our method to edge detection 
and real-valued edge localization.

Sobel gradient operator~\cite{sobel1968} returns the gradient magnitude 
and direction for every image pixel. The high gradient magnitude is an
evidence of a contrast edge. The direction of the gradient is a probable direction
of the edge. We use the output of the gradient operator to define data fidelity
terms of energy~\eqref{eq:energy-w-gamma}. For every pixel $\tilde p_i$ let $g_i$ be the 
gradient vector returned by the operator. We normalize vectors $g_i$ by 
the sample variance of their magnitudes over the whole image. 
We define likelihood $\lambda_i$ using hand picked linear transformation 
of the gradient magnitude: $\lambda_i=1.8-1.4\cdot ||g_i||$.
These parameters were optimized on a single picture shown in Fig.\ref{pic:result-edges}(a).
The initial tangents (line \ref{al:em:init} of Alg.\ref{al:vi}) $l_i$ 
are collinear with gradients $g_i$ and pass through 
 pixels $\tilde p_i$.

\begin{figure}[t]
\begin{tabular}{cc}
\includegraphics[width=0.21\textwidth]{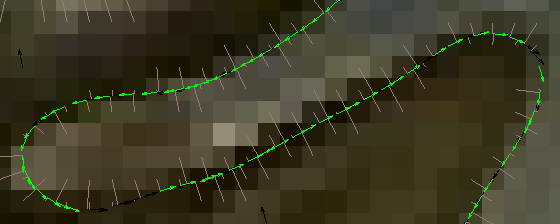} &
\includegraphics[width=0.21\textwidth]{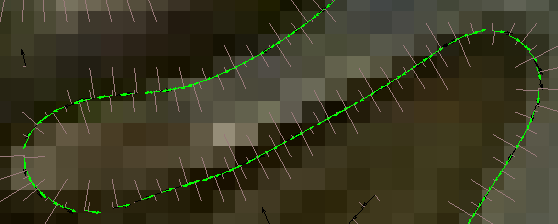}  \\
(a) $\gamma=0$ & (b) $\gamma=0.25$ 
\end{tabular}
\caption{The effect of $\gamma$ in energy~\eqref{eq:energy-w-gamma}. 
Tangents $l_i$ whose $q_i\ge\frac12$ are shown in green, 
tangents such that $\frac14<q_i<\frac12$ are shown in black. 
Increasing $\gamma$ results in increasing probabilities $q_i$ 
of well aligned tangents.}
\label{pic:gamma-effect}
\end{figure}

\begin{figure}[t]
\begin{tabular}{c}
\includegraphics[width=0.15\textwidth]{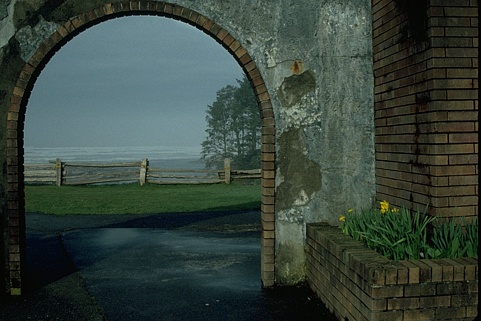}
\includegraphics[width=0.15\textwidth]{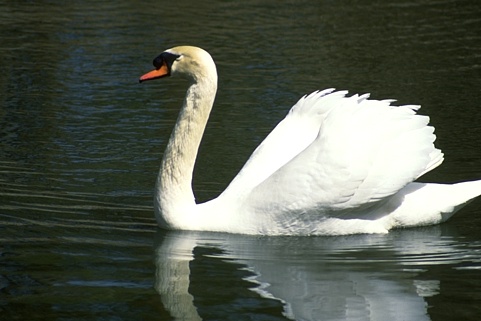}
\includegraphics[width=0.15\textwidth]{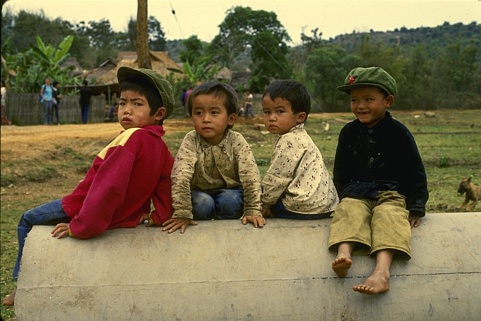}
\\
\includegraphics[width=0.15\textwidth]{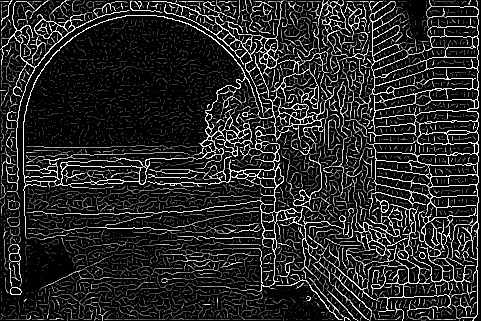}
\includegraphics[width=0.15\textwidth]{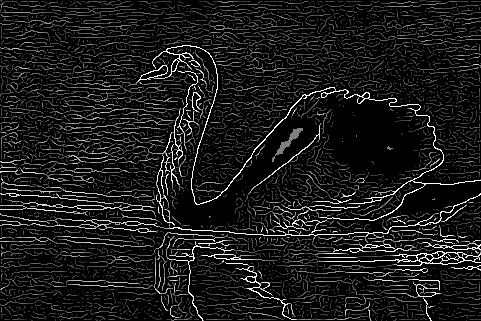}
\includegraphics[width=0.15\textwidth]{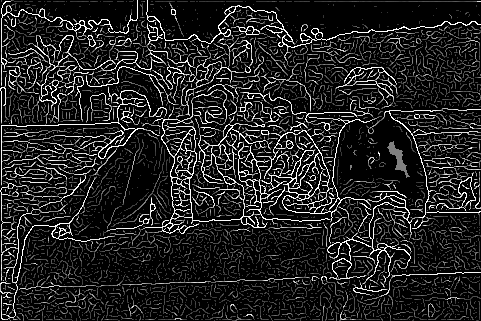}
\end{tabular}
\caption{Examples of the output. The first row shows original images from 
CFGD database\cite{guo2012evaluating}. The second row shows edge
masks at the original resolution produced by our algorithm.}
\label{pic:edge-samples}
\end{figure}

\begin{figure}[t]\centering
\includegraphics[width=0.23\textwidth]{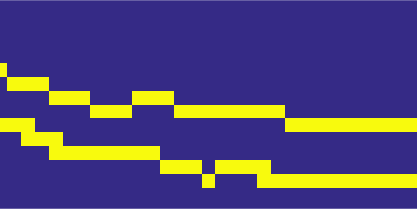}
\includegraphics[width=0.23\textwidth]{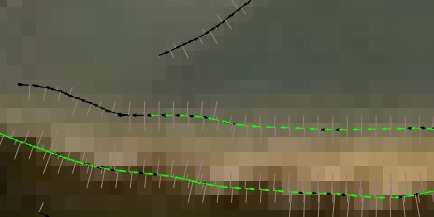}
\caption{Comparison with Canny edge detector~\cite{canny86}. 
Note that Canny only produces the labeling of the pixels.}
\label{pic:us_vs_canny}
\end{figure}

\begin{figure}[ht]
\centering
\includegraphics[width=0.5\textwidth]{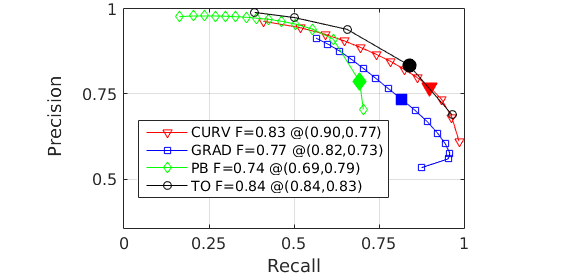} 
\caption{Comparison of out method (CURV) with the baseline gradients (GRAD), 
Pb~\cite{martin2004pb} and the third order filter (TO)~\cite{tamrakar2007no} on 
the database of~\cite{guo2012evaluating}. 
Evaluation of Pb \& TO is given by~\cite{guo2012evaluating}.}
\label{fig:comp}
\end{figure}

The results in figures \ref{fig:sq_vs_abs}, \ref{pic:result-edges}-\ref{fig:comp} 
was obtained by optimizing energy \eqref{eq:energy-w-gamma} 
using \eqref{eq:disc-error} as a soft constraint,
with parameters $\sigma=1$, $\gamma=0.25$ {\NEW and 8-connected neighborhood system $N$}.

According to our model pixel $\tilde p_i$ is a
noisy measurement of point $p$ on a contrast edge. Denoised point $p_i$ is 
the projection of $\tilde p_i$ onto $l_i$.  To generate 
an edge mask (possibly at higher resolution) we can quantize
$p_i$ and use $q_i$ as values at quantized $p_i$.
If during this process we have a conflict such that several points are 
quantized into same pixel we choose the one with maximum probability. 
Fig.\ref{pic:result-edges}(d) shows an edge mask whose resolution was doubled. 
Fig.\ref{pic:edge-samples} shows examples of the edge mask at the original resolution.

We also compared our results with a few edge detection algorithms whose result is an edge mask, 
see Fig.\ref{fig:comp}. This shows that our general method achieves F-measure of $0.83$, 
which is very close to F-measure of $0.84$, given by the best evaluated algorithm in~\cite{tamrakar2007no}.  
Please note that~\cite{tamrakar2007no} was designed specifically for edge detection in images, while our 
approach is a generic method for thin structure delineation.

\subsection{Vessels in 3D}

\begin{figure}[t]\centering
\includegraphics[clip=true,trim=4cm 45mm 2cm 4cm,width=0.5\textwidth]{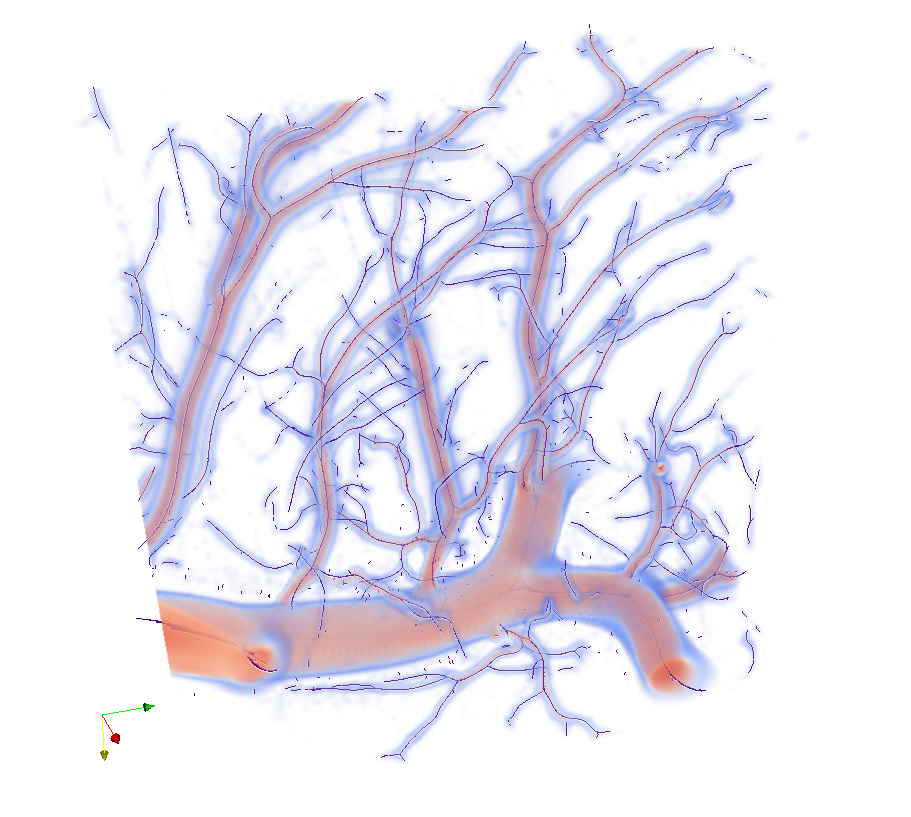}
\caption{Example output of vessel center-line detection in 3D. 
Only tangents $l_i$ with probabilities $q_i\ge\frac12$ are shown (in purple). 
} 
\label{pic:vessels1}
\end{figure}

\begin{figure}[t]\centering
\includegraphics[width=0.474\textwidth]{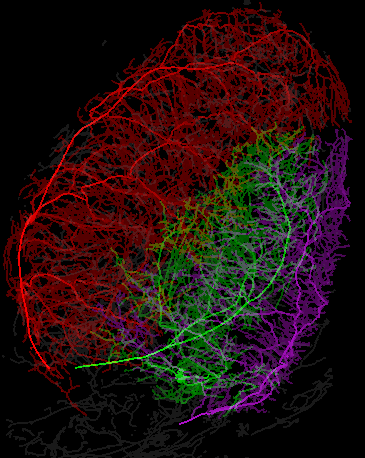}
\caption{Center-line fitting for mouse heart. 
Three main branches are show in color.
Other tangents are shown in dark gray.} \label{pic:vessels2}
\end{figure}

Vessel center-line localization in 3D medical volumes is an important 
task for medical diagnostics and pre-clinical drug trials. 
\fxnote{The sentence looks like an orphan}

For the experiments in this section we used a \emph{microscopic computer tomography}~\cite{granton2008implementation,holdsworth2002micro}
scan of the mouse's heart. The scan is a 3D volume of size 585x525x892. 
For the both experiments the volume was preprocessed with a popular vessel detection filter of~\cite{frangi1998multiscale}.
For every voxel $\tilde p_i$ the filter returns \emph{vesselness measure} $v_i$ such that 
higher values of $v_i$ indicate higher likelihood of vessel presence at $\tilde p_i$. The filter also estimates 
direction $g_i$ and scale $\sigma_i$ of a vessel. 

For this application we use extension~\eqref{eq:energy-w-beta} of energy~\eqref{eq:energy-w-x}.
Coefficient $\frac1\sigma$ in front of the soft constraint in the energy
determines how far tangents $l_i$ can move from voxels $\tilde p$. Since this data has high variability in vessel thickness,  
we cannot use the same $\sigma$ for every voxel. We substitute $\sigma_i$ produced by the vesselness filter
for $\sigma$ in energy~\eqref{eq:energy-w-beta}:
\begin{align*}
E(L,X)= &\ \sum_{(i,j)\in N}  \kappa^2(l_i,l_j)x_ix_j +\\
&+ \sum_{i}\left(\frac{1}{k^2\sigma_i^2}||l_i - \tilde p_i||^2 +\beta{\NEW m}(g_i,l_i) + \lambda_i \right)x_i
\end{align*}
where $k$ is a positive constant and $\lambda_i$ is obtained from vesselness measure $v_i$ by the same linear 
transformation that we use  in Section~\ref{sec:edges}. We set $\beta=0.5$ and $k=20$ {\NEW and use 26-connected neighborhood system $N$}.

For the first experiment we cropped the volume forming a subvolume of size 81x187x173.
We also removed 85\% of voxels with the lowest values of $v_i$. That yields
about $3\cdot10^6$ variables to be optimized. Fig.\ref{pic:vessels1} shows the result.

The goal of the second experiment is to extract a few trees describing the cardiovascular system
of the whole heart. To decrease the running time we perform
Canny's~\cite{canny86} {\NEW hysteresis thresholding to detect one-dimensional ridges} in the volume. 
We substitute vesselness measure for intensity gradients in Canny's procedure. 
Then we set $q_i=1$ for voxels detected as ridges and $q_i=0$ for other voxels.
This yields approximately the same number of optimization variables.
Then we optimize tangents by the algorithm described in Section~\ref{sec:tr}.
Then the estimated center-line points are grouped based on the tangent and proximity information
into a graph and a minimum spanning tree algorithm extracts the trees. The result is shown in Fig.\ref{pic:vessels2}.

\section{Conclusion}\label{sec:conclusion}

We present a novel general early-vision framework for \emph{simultaneous} detection and delineation 
of thin structures with sub-pixel localization and real-valued orientation estimation.
The proposed energy combines likelihoods, indicator (detection) variables and 
squared or absolute curvature regularization. 
We present an algorithm that optimizes localization and orientation variables
considering all possible configuration of indicator variables.
We discuss the properties of the proposed energy and demonstrate 
a wide applicability of the framework on 2D and 3D examples.

In the future, we plan to explore better curvature approximations,
extend our framework to image segmentation with curvature regularization, and improve the running time
by developing parallel GPU implementation.

\section*{Acknowledgements}
{\FIX
We thank Olga Veksler (University of Western Ontario) for insightful discussions.
}

{\small
\bibliographystyle{ieee}

\bibliography{curvature_iccv15}
}
\end{document}